\documentclass[letterpaper, 10 pt, conference]{ieeeconf}  

\IEEEoverridecommandlockouts   
\let\labelindent\relax
\usepackage{enumitem}
\usepackage{times}
\usepackage{amsmath,amssymb}
\usepackage{graphicx}
\usepackage{hyperref}
\usepackage{cleveref}
\usepackage{booktabs}
\usepackage{caption}
\usepackage{color}
\usepackage{pgfplots}
\usepackage[tight]{subfigure}
\usepgfplotslibrary{groupplots}
\usepackage{multirow}
\usepackage{float}
\usepackage{setspace}
\usepackage{tikz}
\usepackage{tikzscale}
\usepackage{subfloat}
\usepackage{wrapfig}
\usepackage{xspace}
\usepackage{cite}
\usepackage{indentfirst}
\usepackage{comment}

\usepackage{titlesec}

\titlespacing*{\subsection}{0pt}{1.0ex}{0.5ex}
\titlespacing*{\subsubsection}{0pt}{1.0ex}{0.5ex}

\newcommand{\changliu}[1]{\normalsize{\color{blue}(\textbf{CL:}\ #1)}}

\title{\LARGE \bf Autonomous Integration and Improvement of Robotic Assembly using Skill Graph Representations}

\author{Peiqi Yu$^{1,*}$,
Philip Huang$^{2,*}$,
Chaitanya Chawla$^{2,*}$, \\
Guanya Shi$^{2}$,
Jiaoyang Li$^{2}$,
and Changliu Liu$^{2}$%
\thanks{$^*$ Equal Contribution}
\thanks{$^1$ Peiqi Yu is with the Department of Electrical and Computer Engineering, Carnegie Mellon University, {\tt\small peiqiy@andrew.cmu.edu}}
\thanks{$^2$ Philip Huang, Chaitanya Chawla, Guanya Shi, Jiaoyang Li, and Changliu Liu is with the Robotics Institute, Carnegie Mellon University, {\tt\small \{yizhouhu, cchawla, guanyas, jiaoyanl, cliu6\}@andrew.cmu.edu}}
}

\date{}

\begin{document}
\maketitle

\begin{abstract}
Robotic assembly systems traditionally require substantial manual engineering effort to integrate new tasks, adapt to new environments, and improve performance over time. This paper presents a framework for \emph{autonomous integration and continuous improvement} of robotic assembly systems based on \textbf{Skill Graph representations}. A Skill Graph organizes robot capabilities as verb-based skills, explicitly linking semantic descriptions (verbs and nouns) with executable policies, pre-conditions, post-conditions, and evaluators. We show how Skill Graphs enable rapid system integration by supporting semantic-level planning over skills, while simultaneously grounding execution through well-defined interfaces to robot controllers and perception modules. After initial deployment, the same Skill Graph structure supports systematic data collection and closed-loop performance improvement, enabling iterative refinement of skills and their composition. We demonstrate how this approach unifies system configuration, execution, evaluation, and learning within a single representation, providing a scalable pathway toward adaptive and reusable robotic assembly systems.  The code is at https://github.com/intelligent-control-lab/AIDF. 
\end{abstract}

\section{\textbf{Introduction}}

Robotic assembly remains a central challenge in industrial automation due to task variability, environmental uncertainty, and the high cost of system integration. Even for well-structured assembly tasks, deploying a robotic system typically involves extensive manual configuration, custom scripting, and iterative tuning by domain experts. As a result, adapting an existing system to new products or improving performance over time is often slow and labor-intensive.

Recent advances in learning-based control and foundation models have shown promise in improving robot autonomy, particularly in perception, manipulation, and decision-making. However, their integration into real-world robotic assembly systems remains challenging for several fundamental reasons. First, many learning-based approaches struggle to reliably generalize to long-horizon, multi-stage assembly tasks, where small errors can accumulate over time and lead to task failure. Second, these methods often rely heavily on large-scale human demonstration data, which may be difficult to obtain for complex industrial setups and, more importantly, may be insufficient for high-precision manipulation tasks that require accuracy beyond human demonstration fidelity.

On the other hand, for conventional non-learning-based systems, there remains a significant gap between high-level task semantics (e.g., ``pick'', ``insert'', ``fasten'') and low-level executable controllers. Task logic is frequently encoded in ad-hoc scripts that tightly couple symbolic intent with platform-specific control implementations, making it difficult to generalize to new tasks, products, or robot platforms. Finally, across both learning-based and classical approaches, there is a lack of explicit structure for systematically evaluating deployed skills, diagnosing failure modes, and driving principled performance improvement over time. As a result, system integration, adaptation, and optimization are often treated as separate and manual processes, rather than as part of a unified autonomous framework.

In this paper, we propose \textbf{Skill Graph representations} as a unifying abstraction for autonomous integration and improvement of robotic assembly systems. The core idea is to represent robot capabilities as a graph of skills, where each skill is defined by:
\begin{enumerate}[label=(\arabic*)]
  \item a semantic description in terms of verbs and associated objects,
  \item executable implementations,
  \item pre-conditions and post-conditions, and
  \item evaluation criteria.
\end{enumerate}

\begin{figure*}[htbp]
    \centering
    \includegraphics[width=\textwidth]{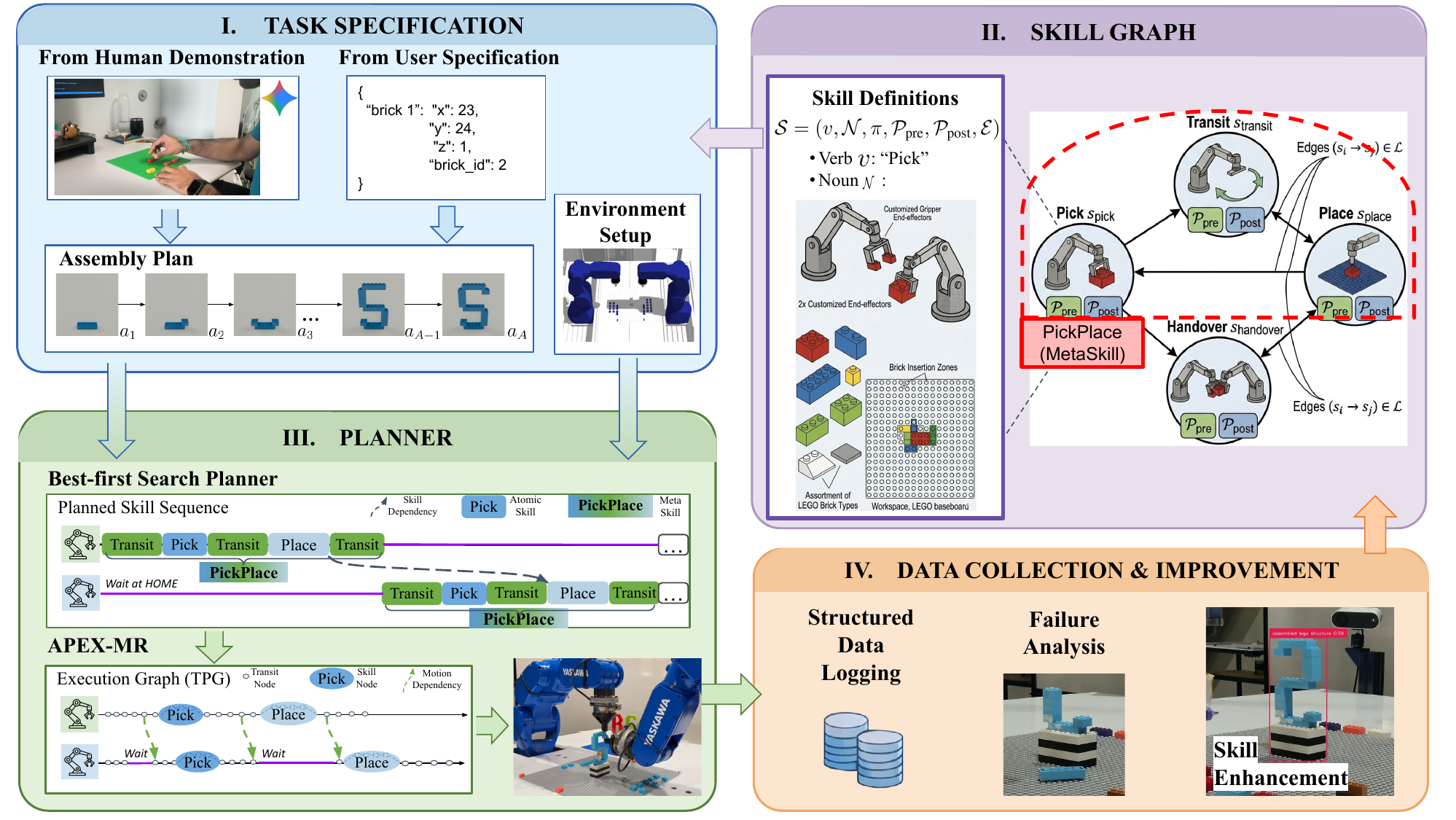}
    \caption{Overview of the Skill Graph representation and its integration with a bimanual robotic LEGO assembly task. }
    \label{fig:overview}
\end{figure*}

As illustrated in Fig. \ref{fig:overview}, we show how this representation enables rapid system integration by allowing assembly tasks to be specified and planned at the semantic level, while preserving precise execution semantics. Furthermore, once deployed, the Skill Graph provides a natural scaffold for data collection and performance-driven improvement, enabling the system to iteratively refine individual skills and their compositions.

\section{\textbf{Related Work}}

\subsection{\textbf{Task and Skill Representations in Robotics}}

Robotic task execution has traditionally been structured using behavior trees, finite-state machines, hierarchical task networks (HTNs) \cite{nau1999shop}, and options in reinforcement learning \cite{Colledanchise_2018, KaelblingLozano‑Perez2013, sutton1999between}. Classical symbolic planners such as PDDL \cite{McDermottEtAl1998PDDL} and Answer Set Programming \cite{erdem2012answer} model tasks as discrete operators with logical preconditions and effects, enabling formal reasoning. Extensions including FastDownward \cite{fastdownward} and HDDL \cite{Höller_Behnke_Bercher_Biundo_Fiorino_Pellier_Alford_2020} improve hierarchical expressivity and search efficiency.

However, purely symbolic approaches struggle in dynamic environments and in bridging symbolic plans with continuous geometric reasoning and control \cite{KOOTBALLY201542, jiang2019task}. While recent work incorporates online action instantiation, concurrency reasoning, and replanning \cite{Buehler_Pagnucco_2014, 10.1007/978-3-031-63227-3_31}, the gap between symbolic specification and executable control remains a central challenge.

\subsection{\textbf{Learning-Based Manipulation and Assembly}}

Imitation learning and reinforcement learning have achieved strong performance in contact-rich manipulation and assembly \cite{levine2016end, qiu2025humanoid, zhu2019dexterous}. Although robust within specific domains, these methods are often task- or platform-specific and lack standardized interfaces for reuse and composition.

Vision-Language-Action (VLA) models \cite{brohan2022rt, brohan2023rt2, octo_2023} integrate perception and language for end-to-end action prediction. Yet many treat complex tasks as monolithic skills, limiting interpretability and compositional generalization when new tasks require recombining learned primitives \cite{mao2025robomatrixskillcentrichierarchicalframework, zhou2025exploringlimitsvisionlanguageactionmanipulations}. These limitations motivate structured and modular skill representations.

\subsection{\textbf{Skill-Centric and Neuro-Symbolic Frameworks}}

Skill-centric and neuro-symbolic approaches decompose tasks into reusable executable components \cite{kroemer, shankar2024translating, mishani2025mosaic}. Control-driven abstractions ground skills in executable controllers rather than purely symbolic operators \cite{ijspeert2013dynamical}, and frameworks such as SCALE \cite{scale} demonstrate improved data efficiency and compositionality through semantically meaningful skills.

Multi-agent extensions such as MA-PDDL \cite{learningrobot} incorporate coordination and spatial constraints but often require manual integration between planning and low-level control, leaving semantic–execution consistency unresolved.

Skill Graph builds upon these works by defining atomic and meta skills as parameterized executors with grounded preconditions, postconditions, and evaluators. Unlike purely symbolic planners, it maintains direct links to controllers; unlike monolithic learning approaches, it supports modular composition and reuse. By integrating semantic structure with execution-aware interfaces, Skill Graph provides an ontology-driven architecture \cite{beetz2018knowrob} that bridges symbolic reasoning and physical grounding for scalable multi-robot manipulation.

\section{\textbf{Skill Graph Representation}}
\label{sec:skill_graph}

\subsection{\textbf{Skill Definition}}\label{subsec:skill_definition}

A \textbf{skill} is defined as a tuple
\begin{equation}
\mathcal{S} = (v, \mathcal{N}, \pi, \mathcal{P}_{\text{pre}}, \mathcal{P}_{\text{post}}, \mathcal{E}),
\end{equation}
where $v$ is a verb representing the action type (e.g., pick, place, insert), $\mathcal{N}$ denotes applicable nouns including objects, robots, and environments, $\pi$ is an executable policy or controller, $\mathcal{P}_{\text{pre}}$ and $\mathcal{P}_{\text{post}}$ are pre-conditions and post-conditions, and $\mathcal{E}$ is a skill evaluator. In each skill, we characterize its dependencies on the applicable objects, robots, and environments, in order to use these as parameters to condition the policy, the pre- and post- conditions, and the evaluator. For example, a ``pick" skill with a parallel gripper will have a different execution policy (which usually grasps on the side of the object) than a ``pick" skill with a suction gripper (which usually sucks on top). But the semantic meaning of these low-level actions remains the same, as the objects are being attached to the robot end-effector, hence would be treated similarly in task description. The type of ``pick'' to choose will be decided by the planner using the skill evaluator $\mathcal{E}$ in the task setting; for big objects that are wider than the gripper width, the skill evaluator will favor the suction gripper. 

Before instantiating the nouns, the skill is still in its abstract form. Once the nouns are determined, e.g., which robot to use, which object to manipulate, in which environment, the skill is concretized and executable. The concretization happens at the task configuration stage, and then a skill graph can be formed to ease planning.

\subsection{\textbf{Skill Graph Structure}}
In a given task setup, for a given set of nouns $\mathcal{N}$, e.g., two robot arms with customized end-effectors, LEGO bricks (as objects to be manipulated), a workspace with LEGO baseboard, we could extract an organized skill graph. 
The skill graph contains a directed graph $\mathcal{G}=(\mathcal{V},\mathcal{L})$, where each node $s\in \mathcal{V}$ corresponds to a skill defined in Sec. \ref{subsec:skill_definition}, and each directed edge $(s_i \rightarrow s_j) \in \mathcal{L}$ encodes feasible transitions between two skills. Transitions in the Skill Graph are governed by the pre-conditions and post-conditions associated with each skill. All conditions are defined over a shared state space $\mathcal{Z}$ as the states of the items in $\mathcal{N}$. We denote the current state as $z \in \mathcal{Z}$, where $z$ can be measured. A directed edge $(s_i \rightarrow s_j)$ is feasible if $\mathcal{P}_\text{post} (s_i, z) \Rightarrow \mathcal{P}_\text{pre}(s_j,z)$, meaning the current execution of skill $s_i$ establishes the conditions required to execute skill $s_j$. Under this formulation, planning over the Skill Graph corresponds to searching for a path of skills, while the execution of the task corresponds to chaining the executables of the skills and ensuring the transitions across skills meet the pre- and post- condition requirements. These aspects will be discussed in the following section. 

\subsection{\textbf{Atomic and Meta Skills}}
Skills are composable. For commonly used skill composition, we can group them together to form a new skill. Here we introduce two concepts, atomic skills and meta skills. 

\textbf{Atomic Skills} correspond to low-level, directly executable actions, such as ``Pick", ``Place", ``Transit", or ``Detect". We define a skill as a parameterized function that either changes the world state or updates the system’s belief about the world state; as a result, both motion skills and perception skills are naturally included in this formulation. In the proposed framework, atomic skills are obtained by instantiating the skill definition $s\in \mathcal{S}$ with concrete executable policies $\pi$ and evaluators $\mathcal{E}$. These policies may be implemented using parameterized motion primitives, wrappers around existing robot controllers, task-specific procedural control logic, or learned black-box policies.

\textbf{Meta Skills} represent higher-level capabilities formed by composing multiple skills into a single reusable abstraction. A meta skill is defined by grouping a sequence or subgraph of skills within the Skill Graph and associating the composition with its own pre-conditions and post-conditions. Internally, a meta skill expands into its constituent atomic skills during execution, while externally it behaves as a single skill node within the Skill Graph. This uniform abstraction allows meta skills to be planned, executed, and evaluated in the same manner as atomic skills. During execution, each atomic skill within a meta skill is instantiated with task-specific parameters.

\section{\textbf{Autonomous System Integration Using Skill Graphs}}
The Skill Graph serves as a central abstraction for integrating planning, execution, and task specification in robotic assembly. By separating semantic definitions from platform-specific implementations, it enables semantic-level planning (Sec.~\ref{subsec:semantic-plan}), execution grounding (Sec.~\ref{subsec:exe-ground}) and intuitive task specification (Sec.~\ref{subsec:intuitive-task-spec}).

\subsection{\textbf{Semantic-Level Planning}}
\label{subsec:semantic-plan}
For robotic assembly, we use a planner to bridge the Skill Graph representation to safe and efficient execution on real robot systems. The key idea is to chain available skills according to their pre- and post-conditions and ground them to specific robots and objects based on task specifications and available resources.

We assume the assembly task is specified as a sequence $\mathcal{A} = \{a_1, a_2, \dots, a_A\}$, and the environment contains $n$ robots and a set of objects $\mathcal{O} = \{o_1, o_2, \dots, o_o\}$. Each task $a_i$ corresponds semantically to a meta skill $s_\text{meta}$ designed to complete that step (e.g., ``assemble'' a brick, which then corresponds to a sequence of atomic skills: ``detect'', ``pick'', ``transit'', ``place''). The planning goal is to ground the assembly sequence to a set of meta skills $s_1, s_2, \dots, s_A$ by assigning a subject (robot) and an  object (to manipulate) to each skill.

Assignments are constrained by the pre-conditions of each meta skill and its associated atomic skills. Skill preconditions may be semantic (e.g., gripper compatibility) or geometric (e.g., kinematic feasibility and reachability). For example, one pre-condition for a pick skill is that there is no object on the robot end-effector and the object to pick is within the reach of the robot. During planning, the state used to evaluate pre- and post- conditions is obtained from simulation. 

We employ a best-first search planner. Each search node represents a partial grounded skill sequence with the evaluator $\mathcal{E}$ assigning cost from simulated execution time. At each step, we pop the top node, enumerate feasible skills that satisfy preconditions for the next assembly step, and insert the resulting nodes back into the queue. The process continues until a fully feasible skill sequence is obtained.

\subsection{\textbf{Execution-Level Grounding}}
\label{subsec:exe-ground}
Executing a planned skill sequence on real robots is challenging due to uncertainties such as controller stochasticity, sensor noise, and environmental variation. During execution, each skill must satisfy its pre- and post-conditions $\mathcal{P}_{\text{pre}}$ and $\mathcal{P}_{\text{post}}$ with respect to the true system states while executing policy $\pi$. Robot-specific parameters (e.g., speed or force limits) can be adjusted to ensure safe and efficient behavior.

To support multi-robot assembly, we provide two execution modes: sequential execution and asynchronous execution based on APEX-MR \cite{huang2025apexmr}.
 
In the sequential execution mode, each meta skill is expanded into atomic skills and executed sequentially, with only one robot active at a time. Intermediate states are propagated via post-conditions to subsequent skills. This mode is safe and simple but inefficient, as no skills are executed concurrently.

In the asynchronous execution (APEX-MR) mode, 
a sequential task and motion plan is postprocessed into a multi-modal Temporal Plan Graph (TPG) \cite{Honig2016-ts} for parallel multi-robot execution. A TPG (e.g., Fig. \ref{fig:overview} Part III.) is a directed acyclic graph where nodes represent robot actions and edges encode precedence constraints. Conflicting actions are ordered explicitly to avoid collisions, and a node can execute only after all incoming dependencies are satisfied. In practice, the policy for each robot skill is parameterized as a trajectory, which may be split into smaller segments corresponding to graph nodes. Intra-robot edges connect consecutive segments, while inter-robot edges enforce collision and task constraints. During execution, a central server dispatches executable nodes to each robot’s action queue and updates dependencies upon completion. Because precedence constraints are encoded in the TPG, concurrent execution remains safe and significantly more efficient than sequential execution.

\begin{figure}[t]
    \centering
    \includegraphics[width=\linewidth]{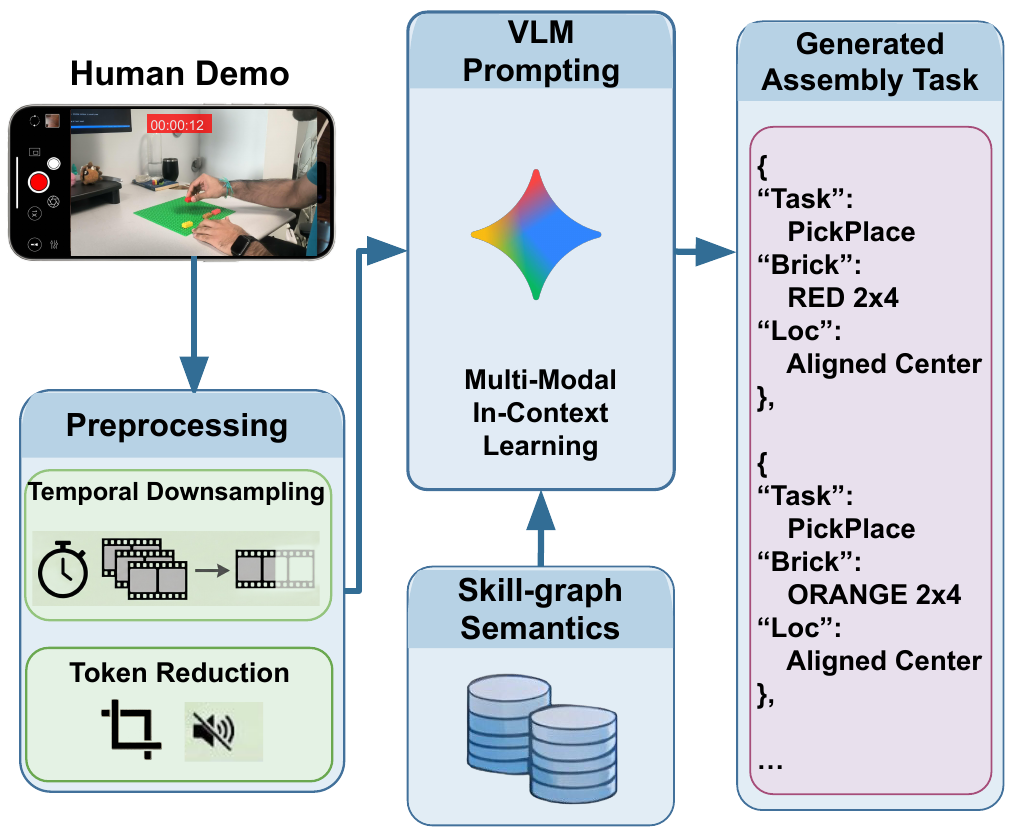}
    \caption{Intuitive Task Specification via Video Extraction. The pipeline transforms a raw human demonstration video into a structured Skill Graph. 
    }
    \label{fig:task_specification}
    \vspace{-5pt}
\end{figure}

\begin{figure*}[htbp]
    \centering
    \includegraphics[width=1.0\linewidth]{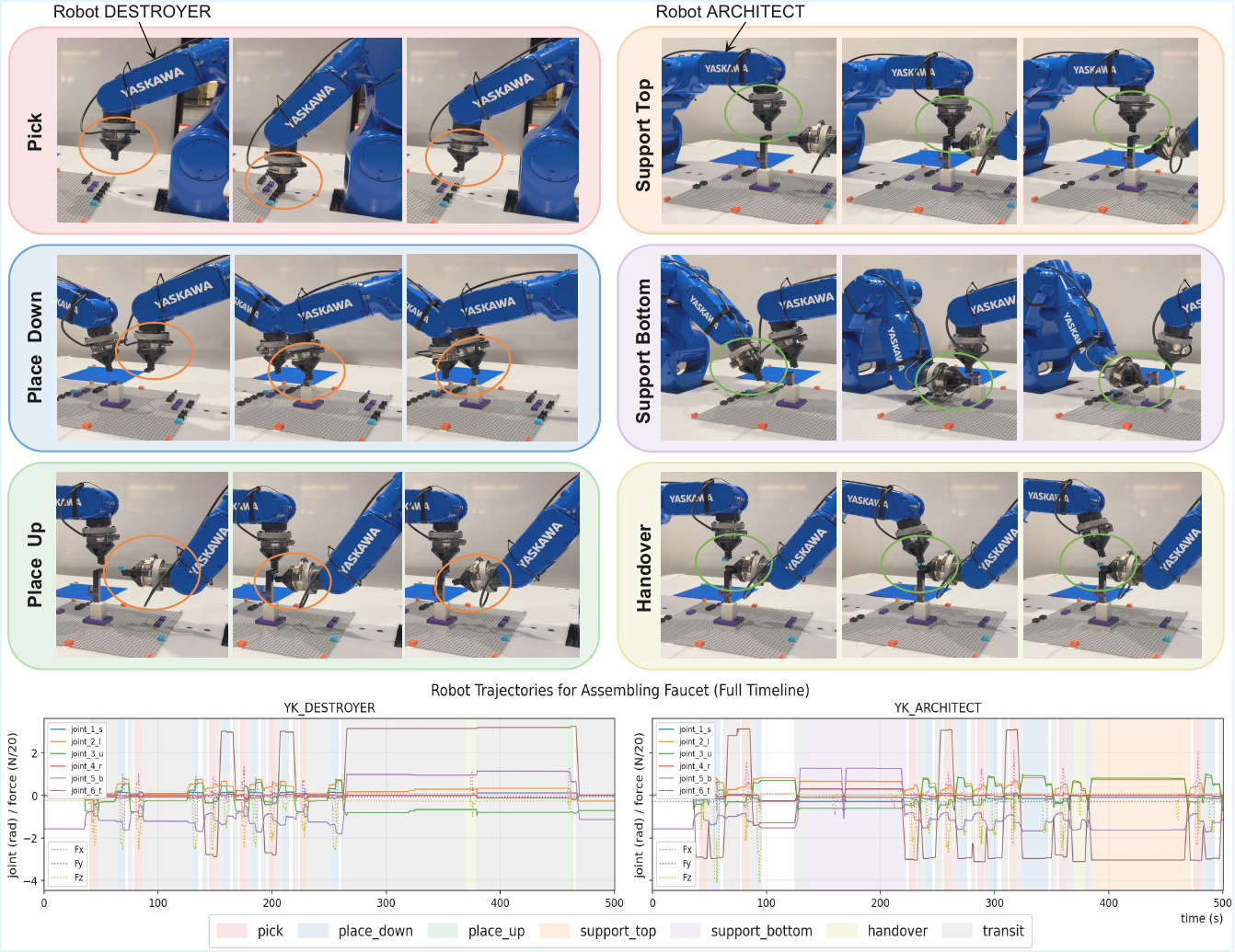}
    \caption{Skill-level trajectory visualization. The upper panels present representative execution snapshots for different skills (Pick, Place Up, Place Down, Support Top/Bottom, Handover, and Transit) performed by the two Yaskawa GP4 robots (named as ``DESTROYER" and ``ARCHITECT"). The lower plots illustrate the temporal evolution of two robots' joint position and force over the full task duration separated by different skill labels, illustrating motion dynamics throughout the entire manipulation sequence with skill labels. }
    \label{fig:skill_traj}
\end{figure*}

\subsection{\textbf{Intuitive Task Specification via Video Extraction}}
\label{subsec:intuitive-task-spec}
\begin{figure}[t]
    \centering
    \includegraphics[width=\linewidth]{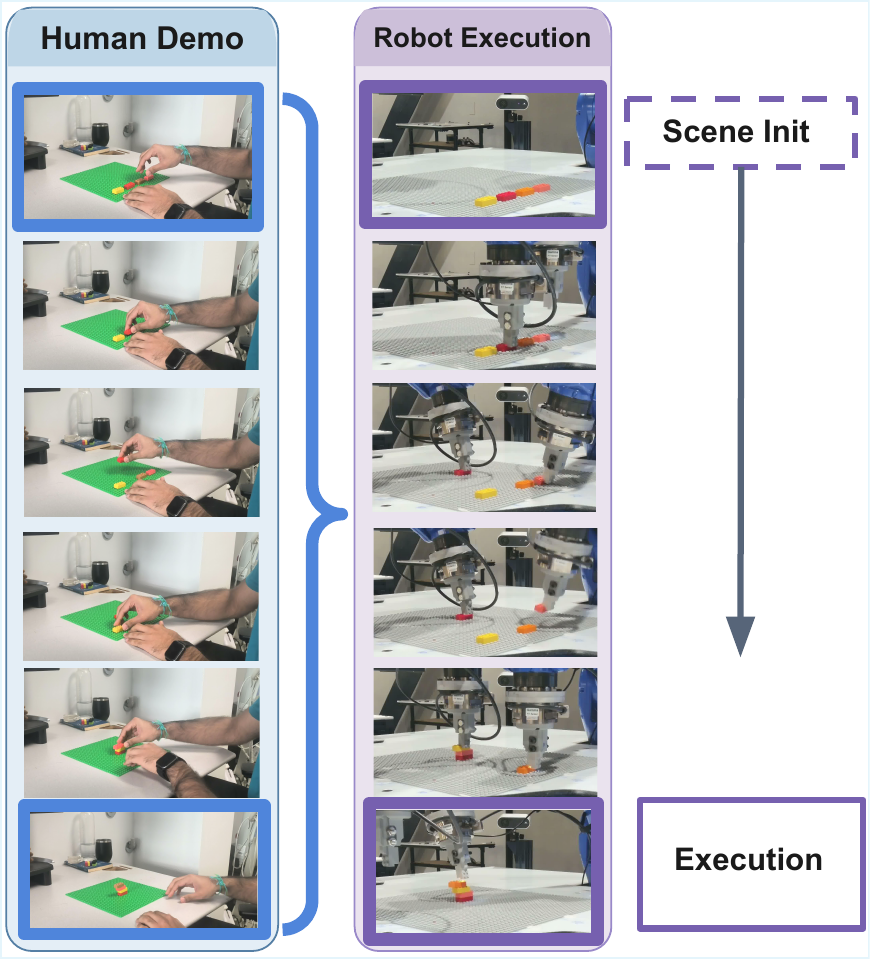}
    \caption{Intuitive Task Specification Results: Transfer from Human demonstration to Robot Execution.}
    \label{fig:human2robot}
    \vspace{-5pt}
\end{figure}

To reduce manual engineering effort in defining task sequences in $\mathcal{A}$, we introduce a video-based pipeline for zero-shot skill sequence extraction from human demonstrations, shown in Fig. \ref{fig:task_specification}. Using a Vision-Language Model (Google Gemini API \cite{team2023gemini}), raw demonstration videos are parsed according to Skill Graph representation $\mathcal{G}$ in Sec. \ref{sec:skill_graph}. The pipeline consists of the following stages. 
1. \textbf{Video Preprocessing and Token Optimization:} The video is temporally downsampled to 10 Hz and cropped to the relevant workspace to reduce token usage and remove visual or audio noise while preserving atomic actions.
2. \textbf{Prompting with Relative Spatial Reasoning:} To improve reliability and reduce hallucinations, we employ a structured prompting strategy. A one-shot multimodal example (image–JSON pair) calibrates the mapping between visual cues and semantic action labels. The initial inventory is injected into the prompt to constrain predictions to physically present objects. Since VLMs cannot reliably regress 3D coordinates, spatial goals are expressed as relative semantic constraints (e.g., “Aligned Center” vs. “Shifted Left”), which are later grounded into physical poses during execution.
3. \textbf{Scene Initialization and State Grounding:} The initial video frames are used to estimate the resource inventory and workspace configuration, defining the initial state $z_0$. The extracted plan is validated against this state to prevent hallucinated actions and ensure resource consistency.
4. \textbf{Schema-Constrained Generation:} The VLM output is restricted to a typed schema defining valid meta skills (e.g., PickPlace, PickPlacewSupport) and object categories according to the skill graph. Structured JSON output ensures direct mapping to the task sequence $\mathcal{A}$.
\textbf{Results:} Fig. \ref{fig:human2robot} shows the results of our pipeline on a human video, feeding the output assembly task to the task-planner, and demonstrating the output trajectory on the real hardware. Note how our pipeline accurately detects the alignment shift in the top two bricks of the structure. For more results, please refer to our supplementary material. 
    


\section{\textbf{Data Collection and Performance Evaluation}}
The Skill Graph not only enables structured execution, but also provides a foundation for systematic data collection and iterative improvement. Each execution generates structured logs that can be reused for analysis, evaluation, and refinement. 
\subsection{\textbf{Skill-Level Logging}}
We pair the Skill Graph with a digital data backbone that records both time-series signals and structured metadata.
During execution, each skill invocation produces robot state trajectories, sensor observations, controller inputs, and evaluations of pre- and post-conditions. 
In addition to streaming time-series data, we store contextual metadata as JSON objects, including skill names, associated entities (robots, objects, tools), policy parameters, and task specifications. This indexed and structured representation enables full reconstruction of skill executions in a digital twin and supports offline analysis at the skill level (Fig.~\ref{fig:skill_traj}). More importantly, these naturally generate structured and labeled data at scale. Such data can be readily leveraged to train end-to-end vision–language–action policies, which we leave for future work.

\subsection{\textbf{Evaluators and Failure Modes}}
The structured logs produced by the Skill Graph provide labeled execution outcomes for each atomic skill. Using these logs, we update skill evaluators $\mathcal{E}$ that estimate the likelihood of successful execution under specific environmental conditions. 
Common failure modes could range from failure to grasp or place a part, an unintended collision between robots and their environment, or a catastrophic structural failure due to accumulated errors from suboptimal assembly steps. 
We train new perception skills using real execution data and later integrate them into the Skill Graph atomic skills or pre/post-conditions to guide skill selection during planning and to monitor execution outcomes online. In this way, the system uses experience collected during deployment to continuously improve the reliability of its skills.

\section{\textbf{Experiments on Robotic Lego Assembly}}

We illustrate the proposed framework on a challenging LEGO assembly task that requires substantial configuration and tuning effort by domain experts.
Starting from a minimal Skill Graph, the system is rapidly integrated and deployed. Through repeated execution, the system collects skill-level data and improves performance, demonstrating increased success rates and reduced execution time.

\subsection{\textbf{System Setup and Skill Instantiation}}

We initialize the Skill Graph with seven (Transit, Pick, Place up, Place Down, Support up, Support Down, Handover) manipulation policies described in \cite{huang2025apexmr} as atomic skills.
The executables of these skills are implemented on two Yaskawa GP4 industrial robots equipped with force-torque sensors, as shown in Fig. \ref{fig:skill_traj}. 
We then form three meta skills - PickPlace, PickPlacewSupport, and PickHandoverPlace from combining these atomic skills. Each of these represents a reusable higher-level capability that can directly manipulate a single LEGO assembly step. 

\subsection{\textbf{Task Specification}}
We define the tasks as a sequence of assembly steps, where each step, denoted as $a_i$, represents a LEGO brick of a specific type (e.g., 2x2 or 2x4) and a discretized position $(x, y, z)$ on a calibrated LEGO assembly plate. 
This assembly sequence can be directly specified by the user or inferred intuitively from a human demonstration video as shown in Fig. \ref{fig:human2robot}. 
In addition, we assume that the current initial state of each LEGO bricks and type are known.

\subsection{\textbf{Semantic Skill Planning and Execution}}
We then plan a grounded skill sequence as described in Sec. \ref{subsec:semantic-plan}. 
For the search-based skill planning, each assembly step is assigned a meta-skill, the corresponding robot arm, and an available LEGO brick. We use the LEGO-specific stability estimator \cite{Liu2024-go}, forward kinematic feasibility, and collision constraints to determine the feasibility of a meta-skill in its pre-condition.
The transit trajectories are planned using RRT-Connect \cite{RRT-Connect}, and we follow Sec. \ref{subsec:exe-ground} to generate the asynchronous execution graph.

\begin{figure}[t]
\centering
\subfigure[Faucet.]{\includegraphics[width=0.24\linewidth]{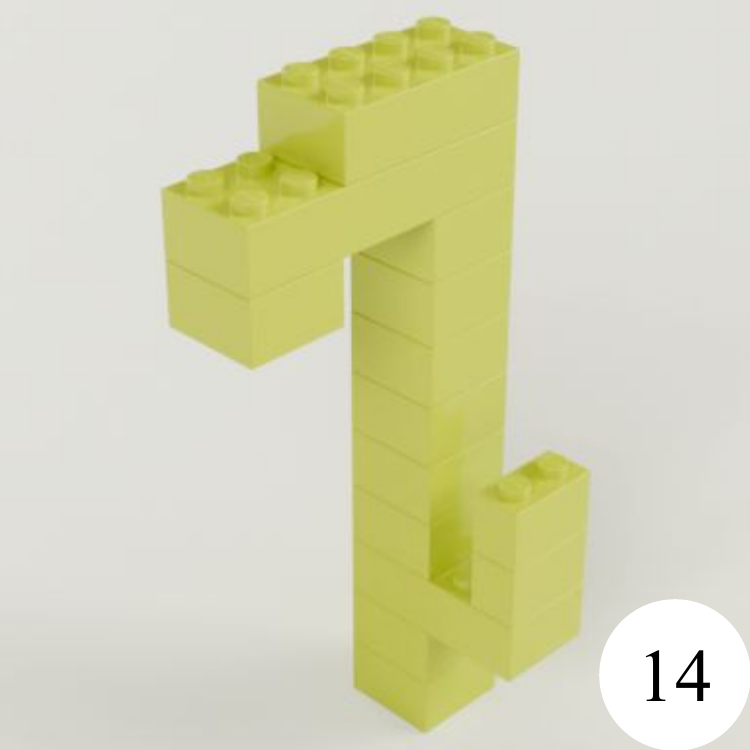}}\label{fig:faucet_render}\hfill
\subfigure[Fish.]{\includegraphics[width=0.24\linewidth]{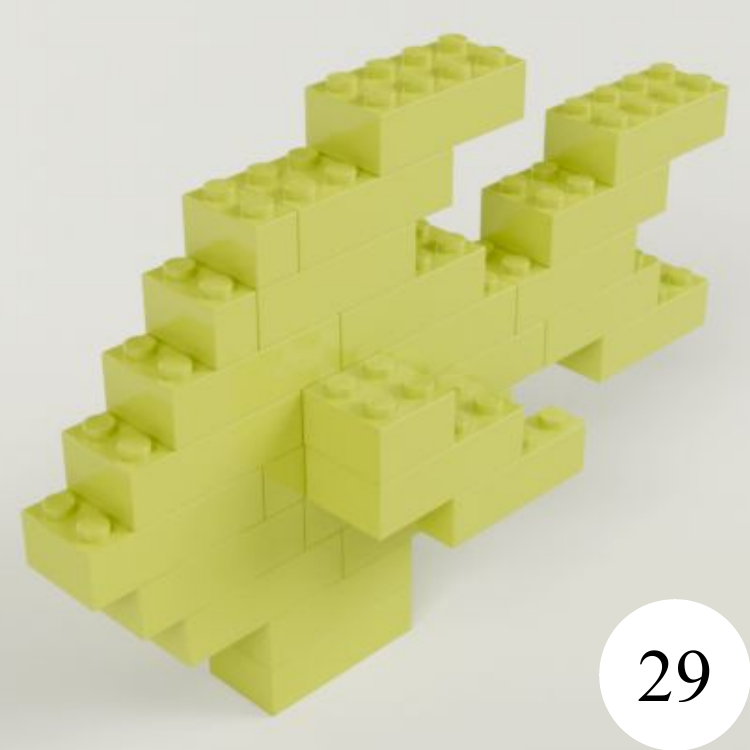}}\label{fig:fish_render}\hfill
\subfigure[Vessel.]{\includegraphics[width=0.24\linewidth]{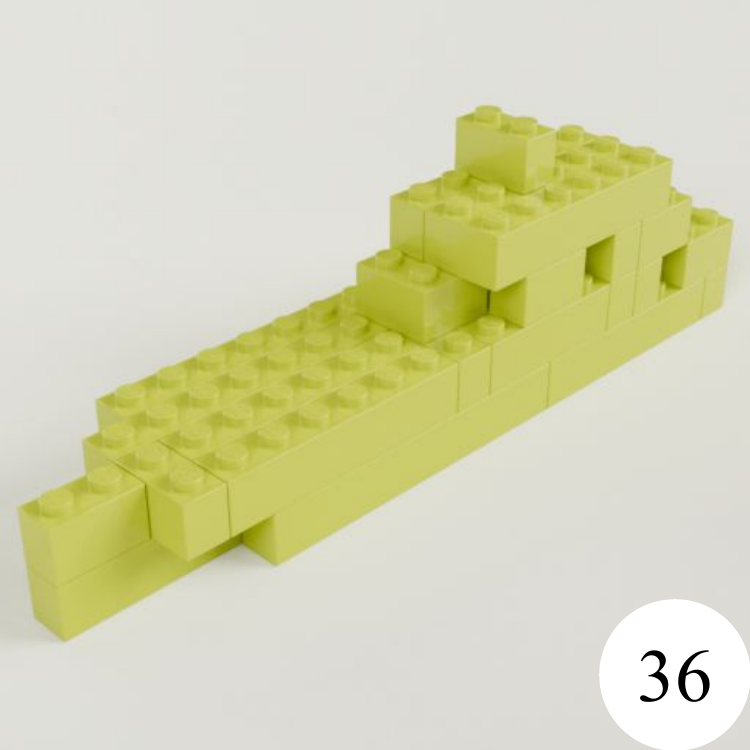}}\label{fig:vessel_render}\hfill
\subfigure[Guitar.]{\includegraphics[width=0.24\linewidth]{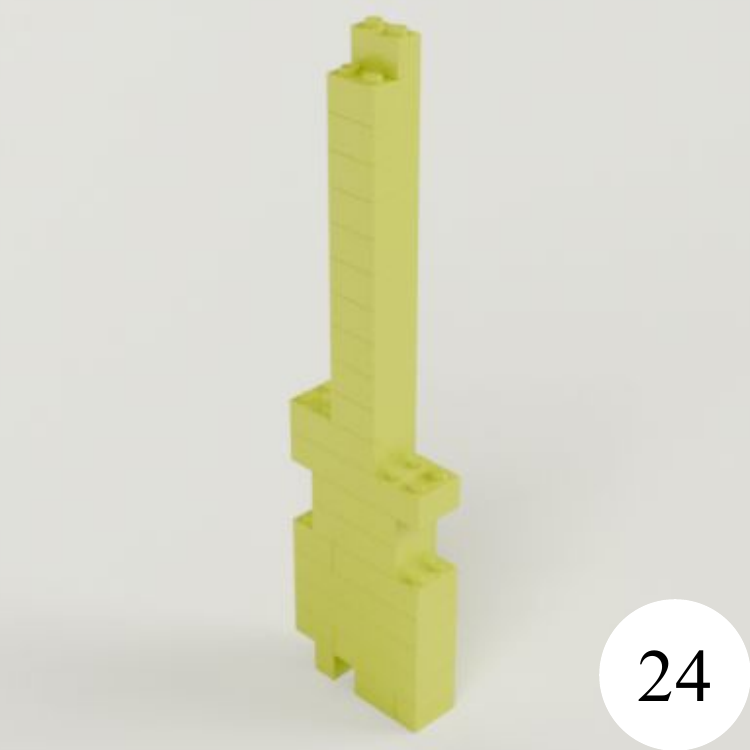}}\label{fig:guitar_render}
\vspace{-5pt}
    \caption{ Brick assembly designs for comparing bimanual assembly construction. The number in each figure indicates the number of bricks required. \label{fig:compare_structures}}
    \vspace{-10pt}
\end{figure}

Fig. \ref{fig:skill_traj} shows an example skill-level trajectory recorded with our data backbone when building a LEGO `Facuet', rendered in Fig. \ref{fig:compare_structures} (a). 
We visualize the temporal evolution of joint states and force readings over the full horizon, as well as video snapshots of all atomic manipulation skills.

\begin{figure}[t]
    \centering
    \includegraphics[width=\linewidth]{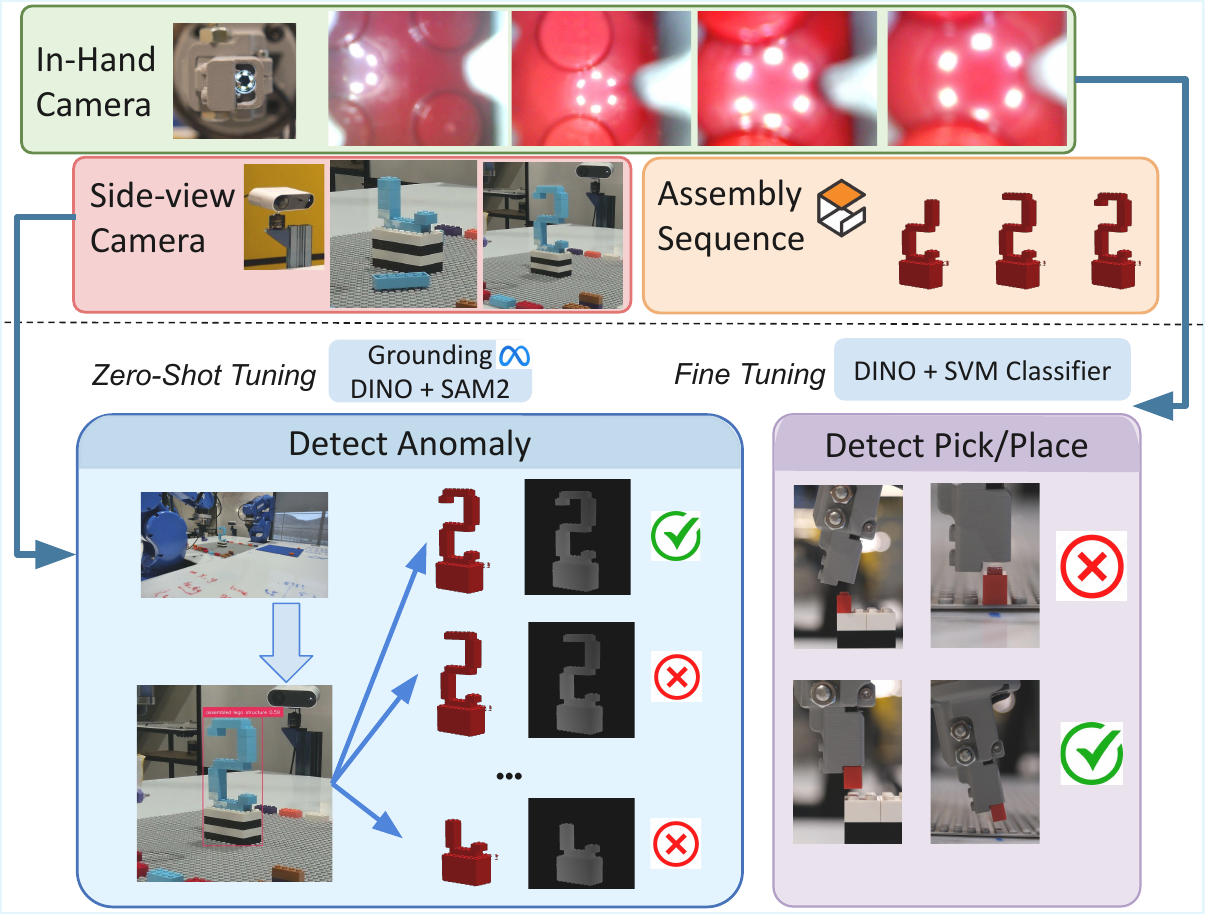}
    \caption{Using planning and execution data to craft new vision-based perception in Skill Graph. In the top, we present the sources of data captured during skill-based planning and execution of a LEGO assembly sequence. The bottom half shows a pick/place post-condition evaluator from an in-hand camera and anomaly detection skill from side-view cameras. }
    \label{fig:fail_det_skills}
    \vspace{-5pt}
\end{figure}

\begin{figure}[t]
\begin{minipage}{\linewidth}
        \centering
        \captionof{table}{ 
        \textbf{Bimanual Construction Comparison.}
        Success Rate: number of trials attempted to have the system successfully build the brick design once without restarting.
        Survival Length: number of bricks (averaged over the attempts) the system assembled without restarting. 
        }
        \label{table:compare_build}
        \setlength{\tabcolsep}{5pt}
\resizebox{\linewidth}{!}{
  \begin{tabular}{lcccc}
    \toprule
    Method & Design & Success Rate & Survival Length \\
    \midrule
    \multirow{4}{*}{Before Skill Improvement} & Faucet & 1/5 & 9.2 
    \\
    & Fish & 0/5 & 7.8 
    \\
    & Vessel & 1/3 & 33.7 
    \\
    & Guitar & \textbf{1/1} & \textbf{24} & 
    \\
    \midrule
    \multirow{4}{*}{After Skill Improvement} & Faucet & \textbf{1/1} & \textbf{14}  
    \\
    & Fish & \textbf{1/1} & \textbf{29} 
    \\
    & Vessel & \textbf{1/1} & \textbf{36} 
    \\
    & Guitar & \textbf{1/1} & \textbf{24} 
    \\
  \bottomrule
\end{tabular}
}

    \end{minipage}
    \vspace{-10pt}
\end{figure}

\subsection{\textbf{Failure Modes and Vision-Based Skill Evaluators}}
Although the manipulation-only Skill Graph can assemble LEGO structures on real robots, execution may fail due to a variety of real-world uncertainties. These issues become more pronounced in long-horizon dual-arm assembly tasks. 
To improve reliability, we analyze the structured execution logs generated during deployment. 
Based on the observed failure modes, we introduce several data-driven improvements to the Skill Graph, including new pre-conditions, post-conditions, and perception skills. These are achieved using additional sensory inputs from an Eye-in-Finger (EiF) camera~\cite{tang2025eye} and multiple third-view cameras (Fig. \ref{fig:fail_det_skills}).

\subsubsection{Enhancing Post-Condition Checks for Pick/Place}
A major source of failure arises from undetected manipulation errors, such as a brick not being grasped successfully or not being released after placement. To address this issue, we use the collected data to train post-condition checkers that verify the outcome of pick and place operations using the EiF camera. A binary classifier built on DINOv2 visual features~\cite{oquab2023dinov2} determines whether the brick is securely in hand after a pick operation or successfully released after a place operation. These checkers allow the system to detect manipulation failures immediately and prevent error propagation to later stages of the task.

\subsubsection{Enhancing Pre-Condition Checks for Pick}
Another common failure mode arises from millimeter-level pose misalignment caused by calibration drift or perception noise. Such errors can lead to failed grasps or unstable placements if left uncorrected. To address this issue, a pre-condition checker is introduced to estimate fine-grained pose offsets using the EiF camera before executing a pick operation. When misalignment is detected, the grasp pose is corrected prior to execution, improving the robustness of the pick skill.

\subsubsection{New Perception Skill for Structural Anomaly Detection}
Long-horizon assembly may also fail due to structural anomalies such as loosened connections, accumulated placement errors, or fragile connections in the LEGO design itself, such as the Fish in Fig. \ref{fig:compare_structures} (b). To detect these situations, a new perception skill is introduced to monitor the partially assembled structure using a third-view camera. The observed structure is compared against the corresponding simulated structure in Gazebo~\cite{koenig2004design}, and a geometric discrepancy measure identifies abnormal assembly states. When anomalies are detected, the system pauses execution and requests corrective intervention from humans.

Together, these improvements extend the manipulation-only Skill Graph into a perception-aware and data-informed framework capable of monitoring manipulation outcomes, correcting geometric errors, detecting structural anomalies, and adaptively allocating verification skills during planning. As shown in Fig. \ref{table:compare_build}, incorporating these improvements significantly increases both task success rate and execution robustness in long-horizon assembly tasks.

\section{\textbf{Discussion}}


In addition to what is shown in the results, we discuss a few ways to further ``close the loop" that utilize the data generated to improve the skill execution,  adapt the task plan, and automate the failure detection and improvement loop. 

\label{sec:skill-improvement}

\subsection{\textbf{Improving Skill Execution}}
\label{subsec:local-improvement}
Collected data could improve execution policies for individual skills through two feedback loops: parametric adaptation for physical consistency and policy adaptation for algorithmic efficiency.
1. \textbf{Parametric Adaptation:}
    Real-world deployments introduce systematic biases due to calibration drift and mechanical wear. To resolve this, we introduce a \textit{Parametric Corrector} that subscribes to logged error signals (e.g., offsets) and maintains an estimate of accumulated bias. Before a skill is executed, it queries this corrector to adjust the target parameters dynamically.  This compensates for drift without requiring manual recalibration or high-level replanning.
2. \textbf{Policy Adaptation}: 
The skill policy supports multiple algorithm implementations (e.g., \textit{RRTConnect} and \textit{BITStar}). Algorithm selection is modeled as a contextual multi-armed bandit. The evaluator could be updated to evaluate the performance of different algorithm implementations using execution feedback (e.g., success rate and runtime). Over time, the planner could learn to select strategies appropriate to task context, such as compliant control for contact-rich tasks and stiff control for free-space motion.

\subsection{\textbf{Failure-Probability-Aware Planning}}
\begin{figure}
\centering
\includegraphics[width=\linewidth]{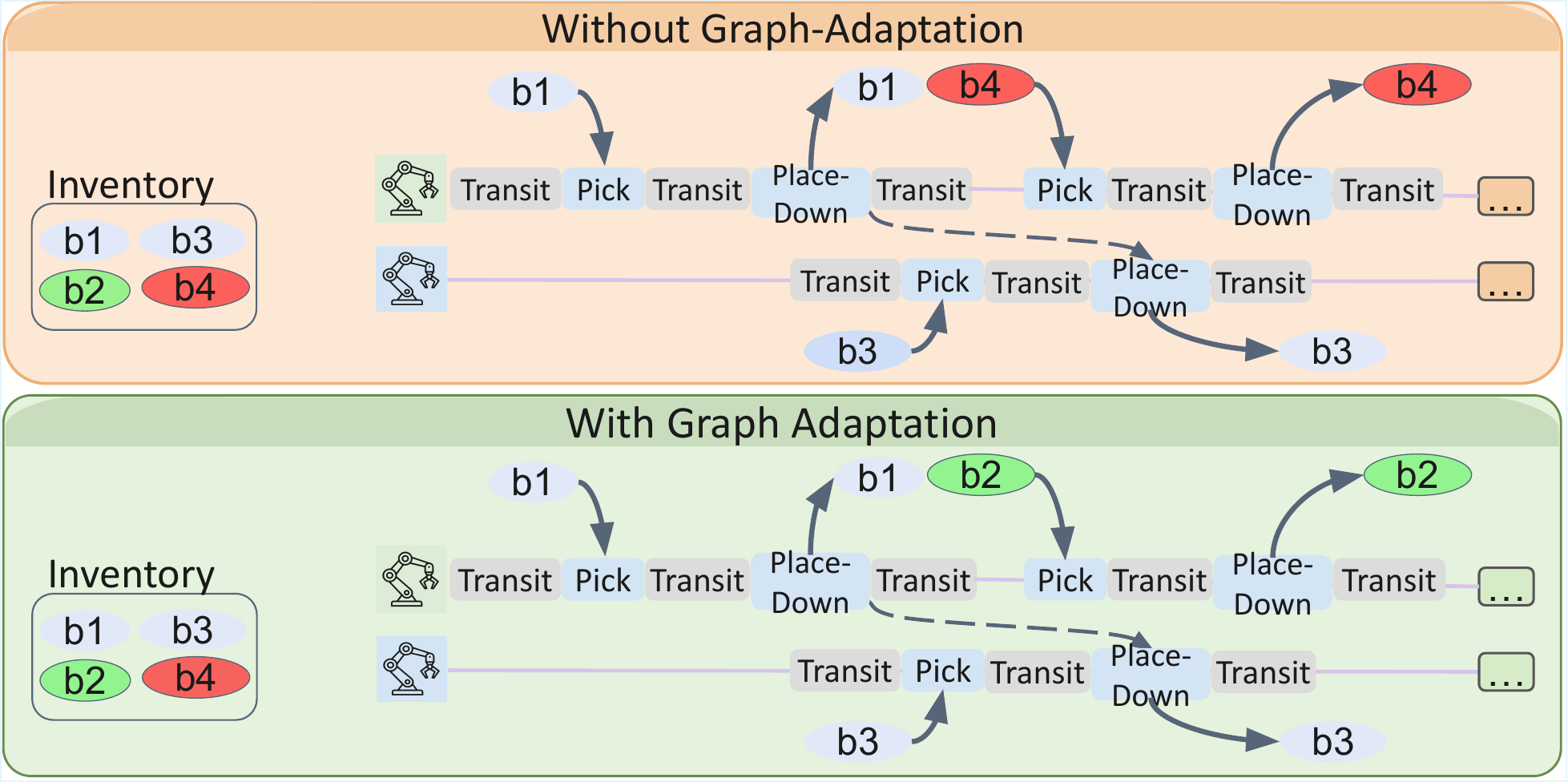}\hfill
\vspace{-5pt}
    \caption{Illustration of graph adaptation. (Top) The planner relies on static costs and selects the high-risk brick b4 (red). \\
    (Bottom) With adaptation, the planner updates the cost tensor with failure history. The solver autonomously reallocates the task to the lower-cost, redundant brick b2 (green), bypassing the failure mode.\label{fig:graph-adaptation}}
    \vspace{-10pt}
\end{figure}

Beyond improving individual skills, execution logs also enable planning-level adaptation. From collected execution traces, the empirical failure probability associated with each skill transition in the Skill Graph is estimated and used to update the corresponding skill evaluators $\mathcal{E}$. These updated evaluators provide failure-probability-aware cost estimates to the planner. During planning, the system could selectively insert perception skills at steps with high estimated failure probability to detect anomalies, while low-risk steps can be executed without additional monitoring. In addition, the planner could also adapt the task allocation to improve the overall success rate, for example by selecting an alternative brick that exhibits less wear (statistically estimated from the failure rate). The adapted graph with an alternative brick is shown in Fig. \ref{fig:graph-adaptation}. Although these adaptations are currently done manually, they are easy to automate. 

\subsection{\textbf{Autonomous Failure Discovery with Foundation Models}}

The Skill Graph representation also enables autonomous improvement through semi-automatic discovery of new failure modes using vision-language models (VLMs). For each execution, structured context is constructed from semantic task descriptions, skill specifications, and execution logs. Logged data, including trajectories, force signals, control commands, and visual observations, are summarized by analysis agents. VLMs extract key visual events from image streams, while signal-level agents compute interpretable temporal features such as force anomalies or trajectory deviations. 
These summaries could be fused with the execution context and analyzed by a language model to infer likely failure causes. As similar patterns recur across executions, the system can identify consistent failure modes and automatically propose new skill adaptations, evaluators, pre- or post-condition checks that capture these conditions. 

\section{\textbf{Conclusion}}

This paper presents a Skill Graph--based framework for autonomous integration and continuous improvement of robotic assembly systems. By structuring robot capabilities around semantic skills with explicit execution and evaluation interfaces, Skill Graphs enable rapid deployment, systematic data collection, and iterative performance improvement. We believe this approach offers a scalable foundation for adaptive, reusable, and safety-aware robotic systems.

\section*{Acknowledgments}

This project is supported by The ARM Institute National Artificial Intelligence Data Foundry for Robotics and the Manufacturing Futures Institute at Carnegie Mellon University.

\bibliographystyle{ieeetr}
\bibliography{references}

\end{document}